\documentclass{article}

\usepackage[nonatbib,preprint]{neurips_2020}

\usepackage[utf8]{inputenc} 
\usepackage[T1]{fontenc}    
\usepackage{hyperref}       
\usepackage{url}            
\usepackage{booktabs}       
\usepackage{amsfonts}       
\usepackage{amsmath,amsthm,amssymb}
\usepackage{scalerel}
\usepackage{mathrsfs}
\usepackage{enumitem}
\usepackage{algorithm}
\usepackage{algorithmic}
\usepackage{mcr}
\usepackage{graphicx}
\usepackage{subcaption}
\usepackage{epstopdf}
\usepackage{todonotes}
\usepackage{multirow}
\usepackage{xcolor}
\usepackage{listings}
\definecolor{codegray}{rgb}{0.5,0.5,0.5}
\lstdefinestyle{Python}{
	language        =   Python,
	basicstyle      =   \ttfamily,
	keywordstyle    =   \color{blue},
	keywordstyle    =   [2] \color{teal},
	stringstyle     =   \color{magenta},
	commentstyle    =   \color{red}\ttfamily,
	breaklines      =   true,
	columns         =   fixed,
	basewidth       =   0.5em,
	numberstyle=\small\color{codegray},
	numbers=left,                    
	numbersep=5pt,  
}
\title{One Backward from Ten Forward, Subsampling for Large-Scale Deep Learning}

\author{
	Chaosheng Dong\thanks{Work done prior to joining Amazon.} \thanks{Equal contribution.}  \\
	Amazon.com Inc.\\
	\texttt{chaosd@amazon.com} \\
	\And
	Xiaojie Jin\textsuperscript{$\dagger$}  \\
	ByteDance Inc.\\
	\texttt{jinxiaojie@bytedance.com} \\
	\And 
	Weihao Gao \\ 
	ByteDance Inc.\\ 
	\texttt{weihao.gao@bytedance.com} \\
	\And 
	Yijia Wang \\
	University of Pittsburgh \\
	\texttt{yiw94@pitt.edu} \\
	\And 
	Hongyi Zhang \\
	ByteDance Inc.\\ 
	\texttt{hongyiz@mit.edu} \\
	\And 
	Xiang Wu \\
	Snap Inc.\\ 
	\texttt{Xwu3@snapchat.com}  \\ 
	\And 
	Jianchao Yang \\ 
	Bytedance Inc. \\  
	\texttt{yangjianchao@bytedance.com} \\
	\And 
	Xiaobing Liu \\ 
	Bytedance Inc.\\
	\texttt{will.liu@bytedance.com} \\
}

\begin{document}
	
	\maketitle
	
	\begin{abstract}
		Deep learning models in large-scale machine learning systems are often continuously trained with enormous data from production environments. The sheer volume of streaming training data poses a significant challenge to real-time training subsystems and ad-hoc sampling is the standard practice. Our key insight is that these deployed ML systems continuously perform forward passes on data instances during inference, but ad-hoc sampling does not take advantage of this substantial computational effort. Therefore, we propose to record a constant amount of information per instance from these forward passes. The extra information measurably improves the selection of which data instances should participate in forward and backward passes. A novel optimization framework is proposed to analyze this problem and we provide an efficient approximation algorithm under the framework of Mini-batch gradient descent as a practical solution. We also demonstrate the effectiveness of our framework and algorithm on several large-scale classification and regression tasks, when compared with competitive baselines widely used in industry.
	\end{abstract}
	
	\section{Introduction}
	Deep neural networks (DNNs) have achieved unprecedented success in many machine learning tasks, for example, in computer vision \cite{krizhevsky2012imagenet,he2016deep}, speech recognition \cite{hinton2012deep}, natural language processing \cite{hochreiter1997long}, recommender systems \cite{covington2016deep}, and game playing \cite{silver2016mastering,silver2017mastering}. As these DNNs typically have a huge number of learnable parameters, they require  millions of data for training. For example, in computer vision, state-of-art DNN models (e.g., \cite{krizhevsky2012imagenet,he2016deep}) use the ImageNet \cite{deng2009imagenet} that contains more than 1.4M images. In natural language processing, BERT \cite{devlin2018bert} uses the BooksCorpus (800M words) and English Wikipedia $(2,500 \mathrm{M}$ words) for the pre-training corpus. In recommendation systems, RecVAE \cite{Ilya2020RecVAE} uses the Netflix dataset that contains more than 100M movie ratings performed by anonymous Netflix customers \cite{bennett2007netflix}. YouTube product-DNN \cite{covington2016deep} uses the dataset that has a vocabulary of 1M videos and 1M search tokens. Moreover, TDM product-DNN \cite{zhu2018learning,zhu2019joint} uses datasets that consist of more than 8M user-book reviews from Amazon and more than 100M records of Taobao user behavior data. With the advent of such large scale datasets, training large DNNs has become exceptionally challenging. For instance, training BERT takes 3 days on 16 TPUv3 \cite{devlin2018bert} and training PlaNet \cite{howard2017mobilenets} even takes 2.5 months on 200 CPU cores using the DistBelief framework \cite{dean2012large}. Thus, there is a growing interest in developing subsampling algorithms to downsize the data volume and accelerate training large DNNs.

	Many approaches have been proposed for data reduction from a wide range of perspectives while preserving the performance as much as possible. Most of these existing approaches fall into one of the two broad categories: Randomized methods and Non-randomized methods. Although both categories construct the samples in a non-uniform data-dependent fasion \cite{mahoney2011randomized}, there is a key difference in the data they operate on. The randomized methods construct the samples by directly operating on a randomized sketch of the input covariate matrix \cite{mahoney2011randomized,drineas2012fast,ma2014statistical}.  In contrast, the non-randomized methods operate on a randomized sketch of both the input covariate matrix and the responses \cite{mcwilliams2014fast,wang2018optimal}. These subsampling approaches have been applied to matrix-related problems in large scale machine learning tasks, e.g., linear regression \cite{drineas2006sampling,drineas2011faster,mcwilliams2014fast,ma2014statistical,ma2015statistical, ma2015leveraging,zhu2015optimal}, logistic regression \cite{fithian2014local,huggins2016coresets,wang2018optimal}, and low-rank matrix approximation \cite{mahoney2009cur,clarkson2017low}. We summarize the literature in Table \ref{table: literature}.
	
	\begin{table}[ht]
		\centering
		\caption{Classification of the research in data subsampling}
		\label{table: literature}
		\begin{tabular}{|c|c|c|c|c|}
			\hline
			& linear regression & logistic regression & low-rank approximation & DNN \\ \hline
			Randomized   &  \cite{mahoney2011randomized,drineas2011faster,ma2014statistical,ma2015statistical, ma2015leveraging,clarkson2017low}                 &                   &         \cite{mahoney2009cur,mahoney2011randomized,clarkson2017low}                      &        \\ \hline
			Non-Randomized &     \cite{drineas2006sampling,mcwilliams2014fast}, ours              &          \cite{fithian2014local,huggins2016coresets,wang2018optimal}, ours           &             ours                  &    ours \\  \hline
		\end{tabular}
	\end{table}

	Despite of these impressive algorithmic achievements in tradition machine learning tasks, none of these work on leveraging or leverage-based sampling demonstrates the capabilities to handle large-scale deep learning tasks. The main reason is that most of these methods are inflexible as they are derived only for specific linear, logistic regression models or low-rank approximations. In this paper, we bridge this gap by proposing a novel subsampling methods on approximating the full data empirical risk under the framework the Mini-batch gradient descent (Mini-batch GD). Our method is motivated by minimizing the discrepancy between the true empirical risk on the entire dataset and the empirical risk on the sampled data.  
	

	\textbf{Our contributions:} In this paper, we have two major contributions for methodological developments in subsampling for solving large-scale deep learning. First, we propose the general optimization framework for data subsampling in any machine learning tasks. This is achieved by minimizing the discrepancy between the true empirical risk when training the model on the entire dataset and the empirical risk when training the model on the sampled data. Second, we develop approximation algorithms through two-step relaxations of the previous optimization problem under the framework of Mini-batch GD. We conduct experiments on the synthetic linear regression experiments that provide
	insight, as well as on the MNIST and ImageNet datasets and show that our method can substantially improve the performance across different tasks given a fixed budget.
	
	\section{Related work} 

	\begin{table}[ht]
		\centering
		\caption{Gradien-based vs Loss-based importance sampling}
		\label{table: gradientVsloss}
		\begin{tabular}{@{}ccc@{}}
			\toprule
			& Convex Program & Deep Learning \\ \midrule
			Gradient Norm &   \cite{bordes2005fast,gopal2016adaptive,zhao2015stochastic,chen2019lsh}             & \cite{alain2015variance,vodrahalli2018all,katharopoulos2018not,johnson2018training}              \\
			Loss     &   \cite{jiang2019accelerating,shah2020choosing}, ours             &   \cite{loshchilov2015online,schroff2015facenet,wang2015unsupervised,shrivastava2016training,simo2015discriminative,schaul2015prioritized,jiang2019accelerating,shah2020choosing}, ours            \\ \bottomrule
		\end{tabular}
	\end{table}
	
	
	\textbf{Importance sampling} These methods are most closely related to our proposed techniques. The key idea behind these methods is to replace the uniform distribution used for sampling with a non-uniform distribution instead. Importance sampling has been used to accelerate the training of DNNs in various applications, such as image classification \cite{loshchilov2015online,johnson2018training}, face recognition \cite{schroff2015facenet}, and object detection \cite{wang2015unsupervised,shrivastava2016training}. Specifically, we consider the Stochastic Gradient Descent (SGD) with importance sampling \cite{zhao2015stochastic,katharopoulos2018not,shah2020choosing} in this paper.

	We summarize the literature in Table \ref{table: gradientVsloss}. Among these paper, \cite{shah2020choosing,jiang2019accelerating} are most related to our work. Similar to ours, both approaches use the loss to construct the sampling distribution. \cite{jiang2019accelerating} prioritizes samples with high loss at each iteration while \cite{shah2020choosing} chooses the sample with lowest loss. However, approaches prioritizing samples with high loss are not robust to outliers while approaches using the samples with low loss often leads to low convergence rate and worse testing performance in practical applications. Although shown to be robust against outliers, the test accuracy of \cite{shah2020choosing} are often inferior to other approaches in practical applications. In contrast, we develop algorithms choosing the subset of samples, the average loss of which best approximates that of the whole batch. We show that our approach achieves better balance in terms of robustness and convergence speed than the existing approaches.
	
	
	\textbf{Coresets selection} Also related to our work is the problem of coresets selection since our algorithm consists of solving the coresets selection problem. This problem aims to select a subset of the full dataset such that the model trained on the selected subset will perform as closely as possible to the model trained on the entire dataset. Originating from computational geometry \cite{agarwal2005geometric}, the idea of coresets selection has been successfully employed to design various machine learning algorithms for, e.g.,  k-Means and k-Medians clustering \cite{badoiu2002approximate,har2004coresets,braverman2019coresets}, SVM \cite{tsang2005core}, SVR \cite{tsang2005coreicml}, and logistic regression \cite{huggins2016coresets}. Most recently, algorithms based on coresets selection have also been proposed for CNN \cite{sener2017active}, GAN \cite{sinha2019small}, and continual learning \cite{borsos2020coresets}. 
	
	Among those approaches that use coresets selection, most similar to ours are the batch active learning (AL) in \cite{sener2017active,pinsler2019bayesian} and the bayesian coresets in \cite{huggins2016coresets,campbell2018bayesian,campbell2019automated} . \cite{sener2017active,pinsler2019bayesian} formulate AL as a coresets selection problem. They choose a subset of unlabeled points to label such that a model learned over the selected subset is expected to yield competitive result over the whole dataset. Our algorithms consider a different setting where all the data is labeled. This is often the case in many real world applications, e.g., recommerder systems.
	\cite{huggins2016coresets,campbell2018bayesian,campbell2019automated} consider constructing Bayesian coresets for the Bayesian statistical models, attempting to select a small subset of the data to approximate the log-likelihood of
	the full data. Differently, we consider the general loss function under any empirical risk minimization framework. We consider our work to be complementary to the Bayesian coresets literature.

	\section{Subsampling for Large-scale deep learning}
	\subsection{Background}
	Let $(\mfx_{i}, y_{i})$ be the $ i$-th input-output pair from the training set, $y=g_{\theta}(\mfx)$ be a Deep Learning model parameterized by the vector $\theta,$ and $\mathcal{L}(\mfx, y)$ be the loss function to be minimized during training. The goal of training is to find
	\begin{align}
		\label{deep learning model}
		\theta^{*}=\underset{\theta}{\arg \min } \frac{1}{N} \sum_{i=1}^{N} \mathcal{L}\left(g_{\theta}(\mfx_i), y_i\right),
	\end{align}
	where $N$ corresponds to the number of examples in the training dataset. 
	
	Gradient descent (GD) is a popular  optimization algorithm to solve \eqref{deep learning model} by updating the parameter $ \theta $ at step $ t $ with the entire training dataset in the following way:
		\begin{align}
		\label{gd}
		\theta_{t+1}=\theta_{t}-\eta\sum\limits_{i=1}^{N}\nabla_{\theta} \mathcal{L}\left(g_{\theta_{t}}\left(\mfx_{i}\right), y_{i}\right),
	\end{align}
	where $ \eta$ is the learning rate fixed across the whole learning period.
	
	Stochastic gradient descent (SGD) \cite{ruder2016overview} is a popular variation of  GD to solve \eqref{deep learning model} due to its high computational efficiency, low memory cost, and convergence guarantee. Specifically, SGD performs a parameter update for the training example $(\mfx_{t}, y_{t})$  at step $ t $ as follows:
	\begin{align}
		\label{sgd}
		\theta_{t+1}=\theta_{t}-\eta_{t} \nabla_{\theta} \mathcal{L}\left(g_{\theta_{t}}\left(\mfx_{t}\right), y_{t}\right),
	\end{align}
	where $ \eta_{t} $ is the learning rate at step $ t $.
	
	Mini-batch gradient descent (Mini-batch GD) is another variation of GD that is intermediate between GD and SGD. It selects a subset of training examples $\left\{\left(\mfx_{i}^{t}, y_{i}^{t}\right)\right\}_{i=1}^{n_{t}}$ sampled from $\left\{\left(\mfx_{i}, y_{i}\right)\right\}_{i=1}^{N}$, where $ 1 < n_{t} < N $ is the sample size at each iteration. Then, the parameter is updated by:
	\begin{align}
		\label{mini-sgd}
		\theta_{t+1}=\theta_{t}-\eta_{t} \sum\limits_{i=1}^{n_{t}}\nabla_{\theta} \mathcal{L}\left(g_{\theta_{t}}\left(\mfx_{i}^{t}\right), y_{i}^{t}\right).
	\end{align}

	In the Mini-batch GD framework, all of $\left\{\left(\mfx_{i}^{t}, y_{i}^{t}\right)\right\}_{i=1}^{n_{t}}$ is used in the parameter updating procedure: the DNN model parameter $\theta$ is being updated using the entire batch data $\left\{\left(\mfx_{i}^{t}, y_{i}^{t}\right)\right\}_{i=1}^{n_{t}}$ at each
	iteration.
	
	This batch contains $n_{t}$ training examples, but potentially, many of them will not contribute much to improving the model because the deep learning model begins to classify these examples accurately, especially redundant examples that are well represented in the dataset. The main question is: 
	
	\textit{Can we only keep a (small) subset of $\left\{\left(\mfx_{i}^{t}, y_{i}^{t}\right)\right\}_{i=1}^{n_{t}}$ for training and throw away the rest? And how should we implement this scheme so that previous performance stays close to intact?}

	\subsection{A general framework for data subsampling}
	
	Before discussing about selecting a subset of $\left\{\left(\mfx_{i}^{t}, y_{i}^{t}\right)\right\}_{i=1}^{n_{t}}$ for training in Mini-batch GD, we propose a general framework for data subsampling for machine learning tasks. Formally, denote $ \mathcal{C} $ the training dataset. Denote $ \mathcal{T} $ the testing dataset. Then, one way to formulate the data subsampling problem is as follows
	
	\begin{align}
		\label{model: data subsampling2}
		\begin{array}{llll}
			\min\limits_{z_{i}, \hat{\theta}_{\mathcal{C}}, \theta^{*}_{\mathcal{C}}} & \norm{\frac{1}{|\mathcal{T}|} \sum\limits_{\left(\mfx_{i}, y_{i}\right) \in \mathcal{T}} \mathcal{L}\left(g_{\theta^{*}_{\mathcal{C}}}(\mfx_{i}), y_{i}\right) - 
				\frac{1}{|\mathcal{T}| }  \sum\limits_{\left(\mfx_{i}, y_{i}\right) \in \mathcal{T}} \mathcal{L}\left(g_{\hat{\theta}_{\mathcal{C}}}(\mfx_{i}), y_{i}\right)} \vspace{1mm}\\ 
			\text { s.t. } & \theta^{*}_{\mathcal{C}} \in \argmin\limits_{\theta} \sum\limits_{\left(\mfx_{i}, y_{i}\right) \in \mathcal{C}} \mathcal{L}\left(g_{\theta}\left(\mfx_{i}\right), y_{i}\right), \vspace{1mm}\\ 
			& \hat{\theta}_{\mathcal{C}} \in \argmin\limits_{\theta} \sum\limits_{\left(\mfx_{i}, y_{i}\right) \in \mathcal{C}}z_{i}\cdot \mathcal{L}\left(g_{\theta}\left(\mfx_{i}\right), y_{i}\right), \vspace{1mm}\\ 
			& \sum\limits_{i=1}^{|\mathcal{C}|} z_{i} \leq K, \vspace{1mm}\\
			& z_{i} \in \{0,1\}, \;\; i = 1, \ldots, |\mathcal{C}|,
		\end{array}
	\end{align}
	where $ K $ is the sample size of the subset we hope to construct. The objective function seeks to measure the discrepancy between the true empirical risk when training the model on the entire dataset and the empirical risk when training the model on the sampled data. The first constraint restricts that $ \theta^{*}_{\mathcal{C}} $ is the optimal estimator when training on the entire dataset. The second constraint restricts that $ \hat{\theta}_{\mathcal{C}} $ is the optimal estimator when training on the sampled data. The third constraint restricts that at most $ K $ data points would be selected during the subsampling process. The last constraint restricts that each data point would be either selected or not selected, where $ z_i = 1 $ means selected, and not, otherwise.

	Solving the formulation \eqref{model: data subsampling2} yields the optimal sampling strategy regarding to achieve the best empirical risk obtainable. However,  note that \eqref{model: data subsampling2} involves many indicator variables and has non-convex objective function and constraints, making it a very complex combinatorial optimization problem and thus NP-hard to solve. Although state-of-art mixed integer non-convex algorithms might solve \eqref{model: data subsampling2} to optimal, it is extremely time-consuming to achieve this and might take much more efforts than solving \eqref{deep learning model} itself. Hence, the general framework is not directly applicable in subsampling the data for large-scale deep learning.


	\subsection{One backward from ten forward for deep learning}
	Due to challenges of solving \eqref{model: data subsampling2} directly, we seek to propose approximation algorithms through two-step relaxations under the framework of Mini-batch GD. 
	
	First, given the data $\left\{\left(\mfx_{i}^{t}, y_{i}^{t}\right)\right\}_{i=1}^{n_{t}}$ at batch $ t $, we approximate the true empirical risk on the whole dataset in \eqref{model: data subsampling2} by the empirical risk on the batch data:
	\[ \frac{1}{|\mathcal{T}|} \sum\limits_{\left(\mfx_{i}, y_{i}\right) \in \mathcal{T}} \mathcal{L}\left(g_{\theta^{*}_{\mathcal{C}}}(\mfx_{i}), y_{i}\right) \approx \frac{1}{n_{t}} \sum\limits_{i=1}^{n_t} \mathcal{L}\left(g_{\theta_{t-1}}(\mfx_{i}^{t}), y_{i}^{t}\right).\]
	
	Second, we approximate the empirical risk on the selected data in \eqref{model: data subsampling2} by the empirical risk on the selected data in batch $ t $:
	\[ \frac{1}{|\mathcal{T}|} \sum\limits_{\left(\mfx_{i}, y_{i}\right) \in \mathcal{T}} \mathcal{L}\left(g_{\hat{\theta}_{\mathcal{C}}}(\mfx_{i}), y_{i}\right) \approx \frac{1}{b} \sum\limits_{i=1}^{n_t} z_{i}^{t} \cdot \mathcal{L}\left(g_{\theta_{t-1}}(\mfx_{i}^{t}), y_{i}^{t}\right),\] where $ b $ is the number of data points we are allowed to sample within a batch.

	Then, we convert the problem of subsampling data points from the entire dataset into the problem of subsampling data points from each batch. Now, one key step of subsampling the training data in the current batch can be formulated as 
	\begin{align}
		\label{model: online_sampling}
		\begin{array}{llll}
			\min\limits_{z_{i}^{t}} & \norm{\frac{1}{n_{t}} \sum\limits_{i=1}^{n_t} \mathcal{L}\left(g_{\theta_{t-1}}(\mfx_{i}^{t}), y_{i}^{t}\right) - \frac{1}{b} \sum\limits_{i=1}^{n_t} z_{i}^{t} \cdot \mathcal{L}\left(g_{\theta_{t-1}}(\mfx_{i}^{t}), y_{i}^{t}\right)} \vspace{1mm}\\ 
			\text { s.t. } & \sum\limits_{i=1}^{n_t} z_{i}^{t} \leq b, \vspace{1mm}\\
			& z_{i}^{t} \in \{0,1\}, \;\; i = 1, \ldots, n_t
		\end{array}
	\end{align}
	where $b$ is the size of the data we hope to sample from  the batch of training data $\left\{\left(x_{i}^{t}, y_{i}^{t}\right)\right\}_{i=1}^{n_{t}}$. \eqref{model: online_sampling} is a sparse subset approximation problem.
	
	Our algorithm for solving large-scale deep learning problem with subsampling of the data points iterates as in Algorithm \ref{alg: online algorithm}.
	\begin{algorithm}[H]
		\caption{One Backward from Ten Forward for deep learning (OBFTF)}
		\label{alg: online algorithm} 
		\begin{algorithmic}[1]
			\STATE {\bfseries Input:} batch data $\{(\mfx_{i}, y_{i})\}_{i=1}^{n_t}$ for $ t = 1, 2, \ldots $, and a budget $ b $.
			\STATE {\bfseries Initialization:} $\theta_{0}$ could be an arbitrary hypothesis of the parameter.
			\FOR{ $t = 1, 2, \ldots$ }
			\STATE \textbf{Forward propagate} and compute $g_{\scaleto{\theta}{4pt}_{\scaleto{t-1}{4pt}}}(\mfx_i^t)$ for $ i = 1,\ldots, n_t $
			\STATE Compute loss $\mathcal{L}\left(g_{\theta_{t-1}}(\mfx_{i}^{t}), y_{i}^{t}\right)$  for $ i = 1,\ldots, n_t $
			\STATE Solve \eqref{model: online_sampling}, get $ z_{i}^{t} $ for $ i = 1, \ldots, n_t $
			\STATE Keep $ (\mfx_{i}, y_{i}) $ if $ z_{i}^{t} = 1 $ 
			\STATE \textbf{Back propagate} and train the model using the selected data, get $ \theta_{t} $
			\ENDFOR
		\end{algorithmic}
	\end{algorithm}
	
	Although \eqref{model: online_sampling} is still a combinatorial optimization problem, it is much easier to solve than \eqref{model: data subsampling2} and there exists efficient approximation algorithms, such as Frank-Wolfe. For the current paper, the combinatorial problem is solved to optimal using state-of-art solver to fully illustrate the performance of Algorithm \ref{alg: online algorithm}. In future, we shall develop fast and accurate algorithms to solve the sparse subset approximation problem.

	\section{Experiments}
	We test the objectives on the following datasets: (1) Synthetic dataset for linear regression, (2) MNIST, and (3) ImageNet. 
	Code is provided in Appendix.
	
	To evaluate the performance of the proposed framework, we compare with the following methods:
	
	\begin{itemize}
		\item \textbf{Uniform Subsampling (Uniform).} Let $\pi_{i}=1 / n,$ i.e., draw the subsample uniformly at random at random.
		\item \textbf{Selective-Backprop.} In each iteration, select the sample with the probability that is proportitial to the current loss \cite{jiang2019accelerating}.
		\item \textbf{Min-$k$ Loss SGD (minK)} \cite{shah2020choosing}. Choosing the subsample with lowest loss.
	\end{itemize}
	
	\subsection{Regression}
	Simulated data: $ y = 2x + 1 + U(-5, 5) $, $ 1000 $ training data, $ 10000 $ testing data.
	\begin{figure}[ht]
		\centering
		\begin{subfigure}[t]{0.4\textwidth}
			\includegraphics[width=1\linewidth]{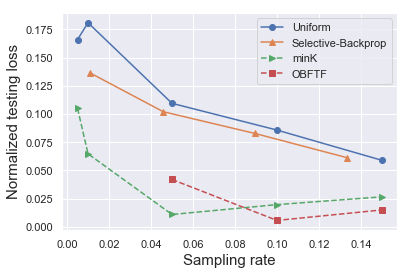}
			\caption{Data without outliers}
			\label{fig:linearRegData}
		\end{subfigure}
		\;\;\;\;\;\;
		\begin{subfigure}[t]{0.4\textwidth}
			\includegraphics[width=1\linewidth]{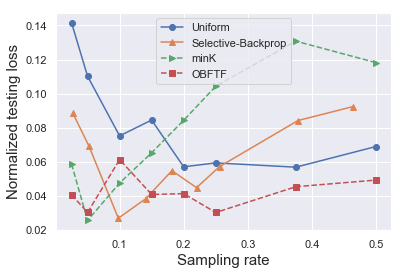}
			\caption{Data with outliers}
			\label{fig:linearRegResult}
		\end{subfigure}
		\caption{Performance of the sampling algorithms for linear regression.}
		\label{fig:linearReg}
	\end{figure}
	
	Simulated data with outlier : $ y = 2x + 1 + U(-5, 5) $ ($ + U(-20, 20) $ for $ 20 $ data points), $ 1000 $ training data, $ 10000 $ testing data.
	
	\textbf{Results}\quad For data without outliers, we train the models with relatively small sampling rate (smaller than 0.15). minK is the most compatitive method in this case. It outperforms other methods with the sampling rate is smaller than 0.05. While OBFTF performs the best when the sampling rate is between 0.1 and 0.15. 
	For data with outliers, we train the models with a variety of sampling rates between 0.01 and 0.5. When the sampling rate is smaller than 0.15, minK and selective-backprop are comparable to OBFTF. Their performance, however, is unstable that a slight increase or decrease in the sampling rate causes a significant drop off in the normalized testing loss. OBFTF performs stable within the sampling rate range, and outperforms other methods when the sampling rate is between 0.15 and 0.5. 
	

	\subsection{Classification: MNIST}
	
	We perform a classification task on
	the MNIST dataset~\cite{lecun1998mnist}, which contains $70,000$ gray scale images of numerical digits from 0 to 9, divided as $60,000$ training images and $10,000$ test images. 
	We do not apply any preprocessing to this dataset and only compare models without data augmentation. 
	
	\begin{figure}[ht]
		\centering
		\includegraphics[width=0.6\linewidth]{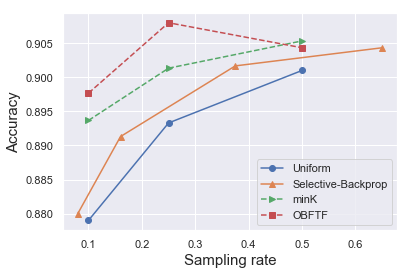}
		\caption{Performance of the sampling algorithms for MNIST.}
		\label{fig:mnistResult}
	\end{figure}
	
	\textbf{Training settings}\quad All the models are trained for 500 epoches with the following settings: initial learning rate is $0.1$, batch size is 128, two hidden layers and both of them have 256 neurons. 
	
	\textbf{Results}\quad We compare OBFTF with the other methods under a variety of sampling ratios. As shown in Figure~\ref{fig:mnistResult}, OBFTF achieves higher accuracy than other methods when the sampling rate is small (0.1 to 0.25), indicating its benefits in speeding up the training phase of classification problems by using a small sampling rate. 
	When the sampling rate increases (to 0.5), the difference of the performance of all these methods are insignificant.
	We note that the accuracy of OBFTF with 0.25 sample rate is higher than the accuracies of all of the methods with 0.5 sample rate! It demonstrates the effectiness of our method in classification tasks, and the fact that a small sampling rate may achieve the same or even better accuracy than a big sampling rate.

	

	\subsection{Classification: ImageNet}
	To further evaluate our method on large-scale datasets,
	we perform a much more challenging image classification task on
	1000-class ImageNet dataset~\cite{deng2009imagenet}, which contains about 1.2 million training images, 50,000
	validation images and 100,000 test images. To verify the effectiveness of our method on different types of neural networks, we choose the popular ResNet50~\cite{he2016deep} and MobileNetV2~\cite{sandler2018mobilenetv2} as baseline models, both of which achieves state-of-the-art results on ImageNet. Compared to ResNet50 which has a higher accuracy, MobileNetV2 is advantageous with higher computational efficiency, thus more friendly to mobile devices, e.g. cell phones.

	\textbf{Training settings}\quad
	Following the training schedule in MNasNet~\cite{tan2019mnasnet}, we train the baseline models using the synchronous training setup on $32$ Tesla-V100-SXM2-16GB GPUs. The initial learning rate is set to be $0.016$, and the overall batch size is $4096$ (128 images per GPU). The learning rate linearly increases to $0.256$ in the first $5$ epochs and then is decayed by $0.97$ every $2.4$ epochs. We use a dropout of $0.2$, a weight decay of $1\mathrm{e-}5$ and Inception image preprocessing~\cite{szegedy2017inception} of size $224 \times 224$. We also use exponential moving average on model weights with a momentum of $0.9999$. All batch normalization layers use a momentum of $0.99$. Using above settings, we train ResNet50/MobileNetV2 for 150/350 epoches respectively.
	
	\textbf{Results}\quad On both ResNet50 and MobileNetV2, we compare our method with the uniform sampling under a variety of sampling ratios, i.e. $[0.1, 0.15, 0.20, 0.25, 0.30, 0.45]$. As can be seen from Table~\ref{tab:imagenet_compare}, on both baseline models, our method achieves higher accuracy than the uniform sampling in this challenging task. Particually when the sampling rate is small e.g. ranging from 0.10 to 0.25, 
	our method has more obvious advantage over the counterpart. This result suggests our method can remarkably benefit training on large-scale datasets by using a small sampling rate to speedup the training phase. When the sampling rate increases, the 
	margin shrinks which may be due to the fact that as sampled out data becomes more representative of the full-sized dataset, the weights of models trained using compared sampling methods receive more accurate gradient updates in each iteration.  Above results demonstrates the effectiness of our method in large-scale machine learning tasks. 
	
	\begin{table}[tb]
		\centering
		\footnotesize
		\caption{Comparison results with the uniform sampling method on ImageNet 2012 \texttt{Val} set. ResNet50 and MobileNetV2 are used as baseline models.}\label{tab:imagenet_compare}
		\begin{tabular}{lccccccc}
			\toprule
			Model & Method & 0.10 & 0.15 & 0.20 & 0.25 & 0.30 & 0.45 \\
			\midrule
			\multirow{3}{*}{ResNet50}& Uniform sampling & 0.7074 & 0.7086 & 0.7316 & 0.7391 & 0.7313 & 0.7430 \\
			& Max prob. & 0.2551 & 0.2986 & 0.3699 & 0.3939 & 0.4431 & 0.4770\\
			& Ours  & 0.7096 & 0.7113 & 0.7355 & 0.7439 & 0.7303 & 0.7452 \\
			\midrule
			\multirow{3}{*}{MobileNetV2}& Uniform sampling & 0.6922 & 0.7065 & 0.7143 & 0.7196 & 0.7242 & 0.7279 \\
			& Max prob. & 0.6164 & 0.6572 & 0.6654 & 0.6700 & 0.6795 & 0.6916 \\
			& Ours  & 0.6956 & 0.7102 & 0.7167 & 0.7198 & 0.7242 & 0.7283 \\
			\bottomrule
		\end{tabular}
		
		\vspace{-1em}
	\end{table}
	
	\section{Conclusion}
	We consider accelerate solving deep learning models in large-scale ML systems. We leverage the key insight that these deployed ML systems continuously perform forward passes on data instances during inference, but ad-hoc sampling does not take advantage of this substantial computational effort. Therefore, we propose to record a constant amount of information per instance from these forward passes. The extra information measurably improves the selection of which data instances should participate in forward and backward passes. A novel optimization framework is proposed to analyze this problem and we provide an efficient approximation algorithm under the framework of Mini-batch GD as a practical solution. We demonstrate the effectiveness of our framework and algorithm on several large-scale classification and regression tasks.

	\bibliographystyle{unsrt}
	\bibliography{reference}
	
	\newpage
	\section*{Appendix: Code}
	\begin{lstlisting}[language=Python]
		import math                                                                                        
		import random                                                                                       
		
		import torch                                                                                        
		import numpy as np                                                                                     
		from ortools.linear_solver import pywraplp  
		
		
		def data_sampling(inputs,
		data_wise_loss,                                                                         
		sampling_ratio,
		sampling_gamma,
		sampling_method):                                                             
		"""A collection of sampling methods used in the paper
		
		Implements our proposed OBFTF and its approximated version, as well as three other competing methods. Tested using Pytorch 1.3.
		
		Args:
		inputs: a torch tensor to store input data.
		data_wise_loss: the loss tensor output by the loss layer, e.g. the cross-entroy loss.
		sampling_ratio: a variable of float representing the ratio of sampled data.
		sampling_gamma: a variable of float, only used in the method of "prob".
		sampling_method: a variable of string, representing the name of method.
		
		Return:
		ind_selected: a 1D pytorch tensor representing the indexes of data which are sampled out.
		"""
		
		ori_input_shape = inputs.shape                                                                             
		if sampling_method == "OBFTF":                                                                                                                                                 
		num_sampled = int(sampling_ratio * inputs.shape[0])
		# refer to the implementation of do_sampling below                                                           
		ind_sign = torch.tensor(                                                                              
		do_sampling(data_wise_loss.detach().cpu().numpy(),                                                               
		inputs.shape[0], num_sampled))                                                                   
		ind_selected = torch.nonzero(ind_sign,                                                                       
		as_tuple=True)[0].long().cuda()                                                      
		
		elif sampling_method == "OBFTF_prox":                                                                      
		num_sampled = int(sampling_ratio * inputs.shape[0])                                                              
		sample_stride = inputs.shape[0] / (num_sampled + 1)                                                                
		ind_sorted = torch.argsort(data_wise_loss, descending=True)                                                            
		ind_sampled = torch.tensor([math.floor(i * sample_stride)                                                             
		for i in range(num_sampled + 1)])[1:].long().cuda()                                                  
		ind_selected = torch.index_select(ind_sorted, 0, ind_sampled).long()                                                        
		
		elif sampling_method == "uniform":                                                                        
		random_thrh = torch.empty(inputs.shape[0],                                                                     
		dtype=torch.float32).uniform_(0, 1)                                                           
		ind_selected = torch.nonzero(random_thrh < sampling_ratio,                                                          
		as_tuple=True)[0]                                                                   
		# guarantee at least one data is sampled out
		if ind_selected.shape[0] == 0:                                                                           
		ind_selected = torch.tensor([random.randrange(0, inputs.shape[0])])                                                      
		ind_selected = ind_selected.long().cuda()                                                                     
		
		elif sampling_method == "prob":                                                                          
		gamma = sampling_gamma                                                                            
		numerator = 1 - torch.exp(-2 * gamma * data_wise_loss)                                                               
		denominator = 1 + torch.exp(-2 * gamma * data_wise_loss)                                                              
		sampling_prob = numerator / denominator                                                                      
		random_thrh = torch.empty(inputs.shape[0],                                                                     
		dtype=torch.float32).uniform_(0, 1).cuda()                                                        
		ind_selected = torch.nonzero(sampling_prob > random_thrh,                                                             
		as_tuple=True)[0].long()                                                                                                                          
		
		elif args.sampling_method == "mink_loss":                                                                       
		def generate_nonrepeat_rand(N, s_n):                                                                        
		ret = set()                                                                                  
		while True:                                                                                  
		if len(ret) == s_n:                                                                            
		return ret                                                                               
		else:                                                                                   
		ret.add(random.randrange(0, N))                                                                    
		
		pool_size = int(sampling_ratio * inputs.shape[0])                                                               
		ind_pool = torch.tensor(list(generate_nonrepeat_rand(     
		inputs.shape[0], pool_size))).long().cuda()                                                                  
		loss_pool = torch.index_select(data_wise_loss, 0, ind_pool)                                                            
		ind_mink = torch.argsort(loss_pool)[0].long()                                                                   
		ind_selected = torch.index_select(ind_pool, 0, ind_mink).long()                                                          
		
		else:                                                                                         
		raise NotImplementedError(sampling_method)                                                                                                                                                                                              
		
		return ind_selected                                                                           
		
		
		def do_sampling(loss, batch_size, sample_num):
		"""The core function in OBTFT: perform sampling by solving an 
		optimization problem
		
		Args:
		loss: the loss of input data in numpy array
		batch_size: the number of input data
		sample_num: the number of required data after sampling
		
		Return:
		a list of binary values, whose length is equals to the the number 
		of input data, i.e. batch_size. 1 means the input data in 
		corresponding index is selected.
		"""
		
		# Create the mip solver with the CBC backend.                                                                     
		solver = pywraplp.Solver('simple_mip_program',                                                                     
		pywraplp.Solver.CBC_MIXED_INTEGER_PROGRAMMING)                                                        
		infinity = solver.infinity()                                                                                                                                                                            
		vars = {}                                                                                       
		for i in range(batch_size):                                                                              
		vars[i] = solver.BoolVar('x[%i]' % i)                                                                                                                                                                     
		var_sub_1 = solver.NumVar(0, infinity, 'var_sub_1')                                                                                                                                                                                                                                                                                                      
		loss_ave = np.random.normal(np.mean(loss), np.std(loss) / np.sqrt(N1), 1)[0]                                                      
		solver.Add(                                                                                      
		loss_ave - solver.Sum([vars[i] * loss[i]                                                                      
		for i in range(batch_size)]) / sample_num <= var_sub_1)                                                       
		solver.Add(                                                                                      
		-loss_ave + solver.Sum([vars[i] * loss[i]                                                                     
		for i in range(batch_size)]) / sample_num <= var_sub_1)
		solver.Add(solver.Sum([vars[i] for i in range(batch_size)]) == N1)                                                                                                                                                                                                                   
		# Minimize the difference                                                                               
		solver.Minimize(var_sub_1)                                                                               
		status = solver.Solve()                                                                                
		
		return [vars[i].solution_value() for i in range(batch_size)]                                                                   
	\end{lstlisting}
\end{document}